\documentclass{article}

\usepackage{arxiv}

\usepackage[utf8]{inputenc} % allow utf-8 input
\usepackage[T1]{fontenc}    % use 8-bit T1 fonts
\usepackage{hyperref}       % hyperlinks
\usepackage{url}            % simple URL typesetting
\usepackage{booktabs}       % professional-quality tables
\usepackage{amsfonts}       % blackboard math symbols
\usepackage{nicefrac}       % compact symbols for 1/2, etc.
\usepackage{microtype}      % microtypography
\usepackage{lipsum}		% Can be removed after putting your text content
\usepackage{graphicx}
\usepackage[numbers]{natbib}
\usepackage{doi}
\usepackage{cleveref}

\title{Evolutionary Computation and Explainable AI: A Roadmap to Understandable Intelligent Systems}

\author{
        Ryan Zhou \\
	Queen's University\\
	\texttt{ryanzhou@queensu.ca} \\
	\And
        Jaume Bacardit \\
	Newcastle University\\
	\texttt{jaume.bacardit@newcastle.ac.uk} \\
	\And
        Alexander Brownlee \\
	University of Stirling\\
	\texttt{alexander.brownlee@stir.ac.uk} \\
	\And
        Stefano Cagnoni \\
	University of Parma\\
	\texttt{cagnoni@ce.unipr.it} \\
        \And
        Martin Fyvie \\
	Robert Gordon University\\
	\texttt{m.fyvie@rgu.ac.uk} \\
        \And
        Giovanni Iacca \\
	University of Trento\\
	\texttt{giovanni.iacca@unitn.it} \\
        \And
        John McCall \\
	Robert Gordon University\\
	\texttt{j.mccall@rgu.ac.uk} \\
        \And
        Niki van Stein \\
	Universiteit Leiden\\
	\texttt{n.van.stein@liacs.leidenuniv.nl} \\
        \And
        David J. Walker \\
	University of Exeter\\
	\texttt{D.J.Walker2@exeter.ac.uk} \\
        \And
        Ting Hu \\
	Queen’s University\\
	\texttt{ting.hu@queensu.ca} \\ 
}

\date{}

\renewcommand{\undertitle}{}
\renewcommand{\shorttitle}{Evolutionary Computation and Explainable AI: A Roadmap}

\hypersetup{
pdftitle={Evolutionary Computation and Explainable AI: A Roadmap to Transparent Intelligent Systems},
pdfsubject={cs.NE},
pdfauthor={Ryan Zhou, Jaume Bacardit, Alexander Brownlee, Stefano Cagnoni, Martin Fyvie, Giovanni Iacca, John McCall, Niki van Stein, David Walker, Ting Hu},
pdfkeywords={Explainability, Interpretability, Evolutionary computation, Machine learning},
}

\begin{document}
\maketitle

\begin{abstract}
Artificial intelligence methods are being increasingly applied across various domains, but their often opaque nature has raised concerns about accountability and trust. In response, the field of explainable AI (XAI) has emerged to address the need for human-understandable AI systems. Evolutionary computation (EC), a family of powerful optimization and learning algorithms, offers significant potential to contribute to XAI, and vice versa. This paper provides an introduction to XAI and reviews current techniques for explaining machine learning models. We then explore how EC can be leveraged in XAI and examine existing XAI approaches that incorporate EC techniques. Furthermore, we discuss the application of XAI principles within EC itself, investigating how these principles can illuminate the behavior and outcomes of EC algorithms, their (automatic) configuration, and the underlying problem landscapes they optimize. Finally, we discuss open challenges in XAI and highlight opportunities for future research at the intersection of XAI and EC. Our goal is to demonstrate EC's suitability for addressing current explainability challenges and to encourage further exploration of these methods, ultimately contributing to the development of more understandable and trustworthy ML models and EC algorithms.
\end{abstract}

% keywords can be removed
% \keywords{Explainability \and Interpretability \and Evolutionary Computation \and Machine Learning}

\section{Introduction}

The proliferation of AI has brought with it an increasing need to understand the reasoning behind its outputs and decisions. While AI methods can learn complex relationships in data and provide solutions to challenging problems, they are often driving decisions that can have significant real-world impacts. The use of predictive models in medicine, hiring, and the justice system has raised concerns about fairness and transparency, and the growing adoption of large language models in commercial products has heightened the importance of avoiding harmful content. Similarly, the application of optimization in areas such as scheduling and logistics~\cite{chen2022evolutionary} requires users to have a robust grasp of the system's operations, as they remain accountable for any adverse outcomes. Consequently, it is crucial not only to improve our models and algorithms, but also to understand and explain the factors driving their prediction or optimization decisions. While active research is ongoing to improve the fairness and safety of AI, this survey focuses on the latter challenge: understanding and explaining AI systems.

Recent AI advancements have heavily relied on ``black-box" approaches. Deep learning, ensemble models, and stochastic optimization algorithms may have well-defined structures, but the processes leading to their decisions are often too complex for human comprehension. In response to this challenge, the field of explainable XI (XAI) has emerged~\cite{iml_murdoch_pnas19}. 

XAI is an umbrella term encompassing research on methods designed to improve human understanding of AI systems' decisions and knowledge capture. It aims to develop techniques that explain AI's decisions, predictions, or recommendations in human-understandable terms. These explanations foster trust, improve system robustness by highlighting potential biases and failures, and provide researchers with insights to better understand, validate, and debug systems effectively. Moreover, they play a pivotal role in ensuring regulatory compliance and enhancing human-machine interactions, allowing users to better discern when they can rely on an AI system's conclusions.

Evolutionary computation (EC) is a powerful approach to AI, with algorithms capable of tackling both optimization and machine learning (ML) tasks. In the context of EC, two directions associated with XAI emerge: first, the application of XAI principles to decision-making within EC, and second, the use of EC to enhance explainability within ML, where the majority of XAI research is currently focused. A growing body of work is developing in both areas, partly fueled by events such as the workshop on EC and XAI held at GECCO in 2022, 2023 and 2024. 

The aim of this paper is to provide a critical review of research conducted at the intersection of EC and XAI. We present a taxonomy of methods and highlight potential avenues for future work, expanding on initial directions proposed in the field of EC~\cite{bacarditIntersectionEvolutionaryComputation2022a, meiExplainableArtificialIntelligence2022, zhouEvolutionaryApproachesExplainable2024}.

The remainder of this paper is structured as follows. \Cref{section:xai} introduces foundational concepts in XAI, such as the nature of explanations and the distinctions between interpretability and explainability, and provides motivation for strengthening the link between XAI and EC. \Cref{sec:ec_for_xai} discusses how EC can be used \textit{for} XAI. \Cref{sec:xai_for_ec} examines how XAI can be \textit{applied to} EC. \Cref{section:outlook} addresses ongoing challenges and potential opportunities. \Cref{sec:conclusion} provides our final thoughts and conclusions.
% \IEEEpubidadjcol

%%%%%%%%%%%%%%%%%%%%%%%%%%%%%%%%%%%%%%%%%%%%%%%%%%%%%%%%%%%%%

\section{Explainable AI}
\label{section:xai}

XAI aims to improve the \textit{understandability} of AI systems -- the degree to which humans can comprehend how a system makes decisions, the reasoning behind those decisions, and their potential implications. Importantly, understandability is subjective and can vary from user to user, depending on their background, experience, and familiarity with the AI system.

XAI employs two general approaches: 1) designing algorithms or models that are easier to understand without external aids, and 2) providing explanations which aid understanding by illuminating an AI system's output process, highlighting significant features and interactions, or revealing potential issues. Even if an AI system is too complex for direct human comprehension, it can be considered \textit{explainable} if it can be understood with the help of these explanations.

Explainability is crucial for several reasons:

\begin{itemize}
\item {\bf Trust}: Explainability directly influences users' willingness to adopt and rely on AI results~\cite{miller_are_2022}. For ML models, it allows users to understand the decision-making process. For optimization, it demonstrates why obtained solutions are reliable.
\item {\bf Validity}: Explanations can reveal whether a solution truly solves the problem or merely exploits an error in the problem definition or a spurious data relationship. This helps avoid surprising or frustratingly incorrect results~\cite{Lehman2020}.
\item {\bf Real-world applicability}: Explanations can reveal important characteristics for optimality, allowing refinement of solutions to better fit real-world problems. This is particularly useful when subtle rules or preferences are difficult to codify in the initial problem definition.
\item {\bf Regulatory compliance}: As AI legislation increases, explanations may provide necessary audit trails for implemented solutions.
\item {\bf Bias detection}: Explanations can help identify unwanted biases in ML predictions, especially when goals like ``fairness" are not explicitly coded in the training cost function.
\end{itemize}

%%%%%%%%%%%%%%%%%%%%%%%%%%%%%%%%%%%%%%%%%%%%%%%%%%%%%%%%%%%%%

\subsection{What is an explanation?}

Defining what constitutes an explanation is challenging. Informally, an explanation aims to answer the question: ``why?''. Prior work has framed explanations in various ways, including providing causal information~\cite{miller2019explanation}, non-causal explanations~\cite{ginet2008defense}, or deductive arguments~\cite{veatchCarlHempelAspects1970}. In this paper, we define an \textit{explanation} as a tool to help humans understand certain aspects of a model or an algorithm. The ultimate purpose of an explanation is to serve as an interface between the model/algorithm and the user, delivering information in a more accessible form. Importantly, an explanation need not capture the full behavior of the model/algorithm but should communicate important insights about it.

Such insights may include the answers to questions such as~\cite{bacarditIntersectionEvolutionaryComputation2022a}:
\begin{itemize}
\item Is the model solving the correct problem, and has the problem been formulated correctly?
\item What are the patterns the model uses for predictions, and are they as expected?
\item Why did the model make this prediction instead of another, and what would change its prediction?
\item Is the model biased, and are its decisions fair?
\end{itemize}

Explanations can take on multiple forms, including visualizations, numerical outputs, data instances, or text descriptions~\cite{iml_murdoch_pnas19}. They may also be part of an ongoing dialogue between a human and an explainer~\cite{miller2019explanation, slackExplainingMachineLearning2023}.

%%%%%%%%%%%%%%%%%%%%%%%%%%%%%%%%%%%%%%%%%%%%%%%%%%%%%%%%%%%%%

\subsection{Explainability and interpretability}

The terms interpretability and explainability are often used interchangeably, but we distinguish them as related yet distinct aspects of understanding a model~\cite{liptonMythosModelInterpretability2018,iML_Rudin_NatML_19}. Similar to understandability, both explainability and interpretability are subjective and depend on the user's knowledge and experience.

Interpretability refers to a human’s ability to follow a model’s decision-making process without external aids. Simple models, like small decision trees or symbolic representations, are considered interpretable. However, as models grow in size or complexity (e.g., random forests, neural networks), they become harder to follow, aligning with the notion that interpretability exists on a spectrum~\cite{liptonMythosModelInterpretability2018}. Simpler models may sacrifice accuracy for ease of understanding, while more complex ones, though more accurate, are less interpretable.

Explainability, on the other hand, refers to the ability to provide human-understandable insights into a model's decisions, even if the exact logic is too complex to trace. Explanations do not need to capture the model’s full behavior; instead, they offer glimpses into how it works, using methods like feature importance, local approximations, or input comparisons.

\begin{figure}[t!]
 \centering
 \includegraphics[width=\linewidth,trim=5mm 50mm 30mm 35mm,clip]{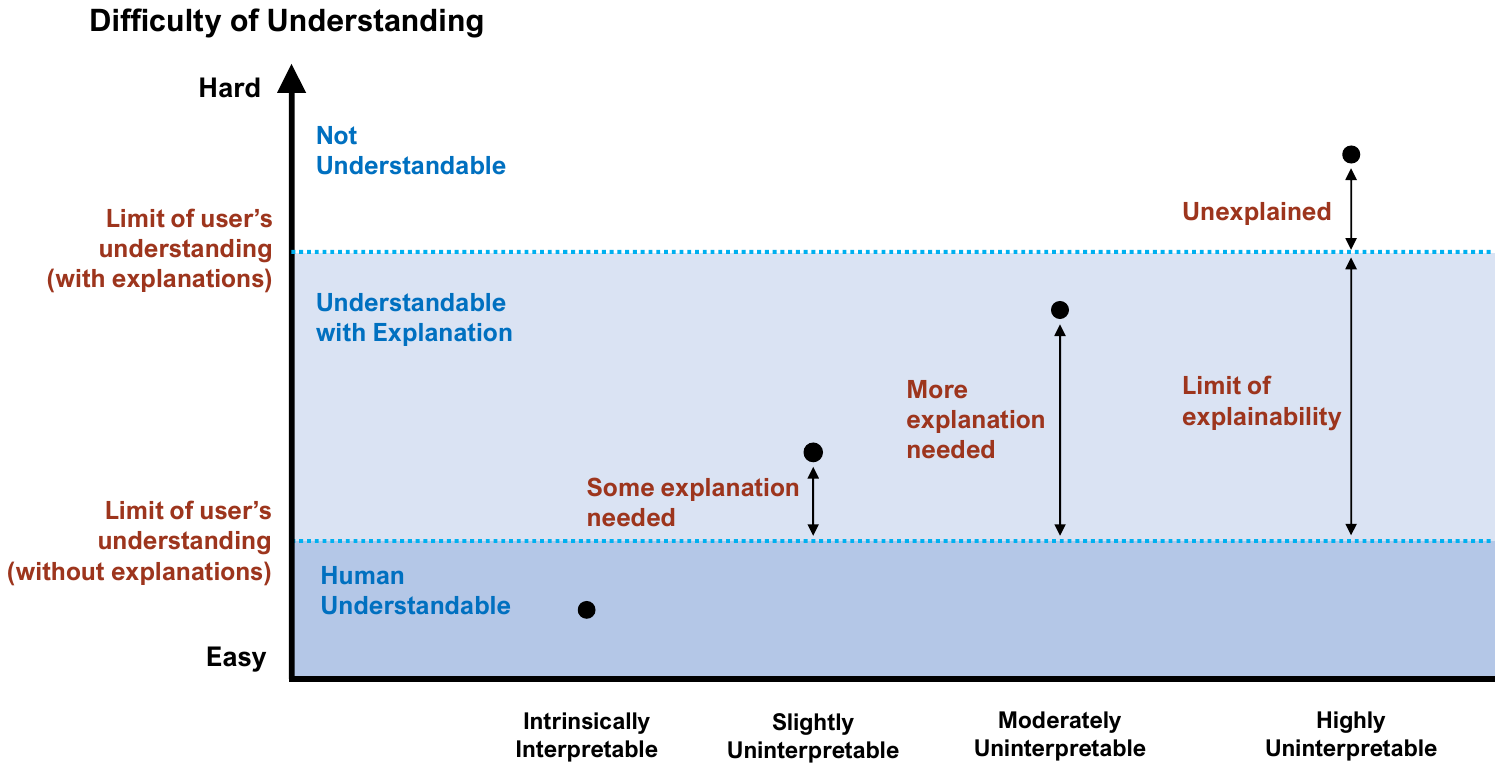}
 \caption{As models and solutions become more difficult to understand, the amount of explanation required increases. Simple solutions (left point) may not require any explanation at all, and are intrinsically interpretable. Others (middle two points) may lie beyond the ability of a human to grasp easily, but can be understood with explanation. Finally, some models (right point) may remain incomprehensible even with the current best efforts at explanation.}
 \label{fig:explainability}
\end{figure}

As shown in Fig.~\ref{fig:explainability}, the more complex a system, the greater the effort required to understand it. Below a certain threshold, models are intrinsically interpretable. Beyond that, explanations are needed to aid understanding. For example, multidimensional models may become understandable with visual aids or feature importance metrics. As the complexity increases further, at some point it becomes impractical to explain the model fully. Explanations of large language models, for example, may still leave key behaviors unexplained. However, this points towards two ways of achieving explainability, which can work in tandem: make the model simpler to understand, or improve our explanation techniques.

%%%%%%%%%%%%%%%%%%%%%%%%%%%%%%%%%%%%%%%%%%%%%%%%%%%%%%%%%%%%%

\subsection{Why EC and XAI?}

Evolutionary Computation (EC) is an AI approach inspired by biological evolution, with applications in optimization, machine learning, engineering design, and artificial life. EC encompasses evolutionary algorithms such as genetic algorithms (GA), genetic programming (GP), and evolution strategies (ES), and extends to swarm intelligence algorithms like particle swarm optimization. These techniques typically use populations of solutions and operators that introduce variation and diversity to explore large regions of the search space.

EC techniques possess unique strengths that can address current challenges in XAI~\cite{arrieta2020explainable, huGeneticProgrammingInterpretable2023}. First, as detailed in later sections, EC has a proven track record of creating symbolic or interpretable models (e.g., decision trees or rule-based systems). By constructing solutions from intrinsically interpretable components, EC-derived solutions can ensure the interpretability of the resulting models. Additionally, EC can generate interpretable approximations of more complex models, producing explanations for their behavior.

Second, the inherent flexibility of evolutionary methods, such as their ability to perform derivative-free, black-box optimization~\footnote{Black-box optimization refers to methods that handle problems where the internal structure, equations, or derivatives of the objective function are unknown or inaccessible. In this context, ``black-box" means the optimizer can work with just the input-output pairs, which is distinct from the use of ``black-box" to describe complex, incomprehensible machine learning models.}, makes them versatile tools for scenarios where other methods struggle. For instance, EC can optimize models accessible only through APIs that provide predictions without revealing internal logic. This flexibility also enables EC to handle customized metrics, such as interpretability metrics, that are not easily optimized via gradient descent. Furthermore, EC can be combined with other algorithms to create hybrid methods or meta-optimizers.

A particularly valuable feature of EC is multi-objective optimization, crucial in XAI where there is often a trade-off between model accuracy and human interpretability or complexity of the explanation. EC can balance these objectives, and by leveraging diversity metrics or quality-diversity algorithms, it can generate a variety of explanations tailored to different users or aspects of the model.

Conversely, XAI approaches can offer valuable insights into evolutionary algorithms and are currently underutilized in EC. XAI can help explain the decision-making process of EC algorithms, making it easier to debug and refine them. This is especially important in fields like engineering design or policy-making, where the rationale behind a solution must be understandable to non-technical decision-makers.

Finally, XAI can enhance the interpretability of fitness landscape analyses in EC. Understanding the fitness landscape is critical for assessing the difficulty of finding optimal solutions and the effectiveness of EC algorithms. XAI-inspired visualization and interpretation tools can provide deeper insights into these landscapes, serving as explanations in their own right.

%%%%%%%%%%%%%%%%%%%%%%%%%%%%%%%%%%%%%%%%%%%%%%%%%%%%%%%%%%%%%
\section{EC for XAI}
\label{sec:ec_for_xai}

This section covers XAI methods for machine learning (ML), and the incorporation of evolutionary algorithms into such methods. As ML models have become more advanced, their complexity has increased, often boosting performance but at the cost of interpretability. Improving explainability is essential to balance this trade-off, ensuring models are not only effective but remain understandable.

%%%%%%%%%%%%%%%%%%%%%%%%%%%%%%%%%%%%%%%%%%%%%%%%%%%%%%%%%%%%%

\subsection{Explainability and complexity}
\label{section:complexity}

\begin{figure}[t!]
 \centering
 \includegraphics[width=0.60\linewidth,trim=25mm 38mm 40mm 35mm,clip]{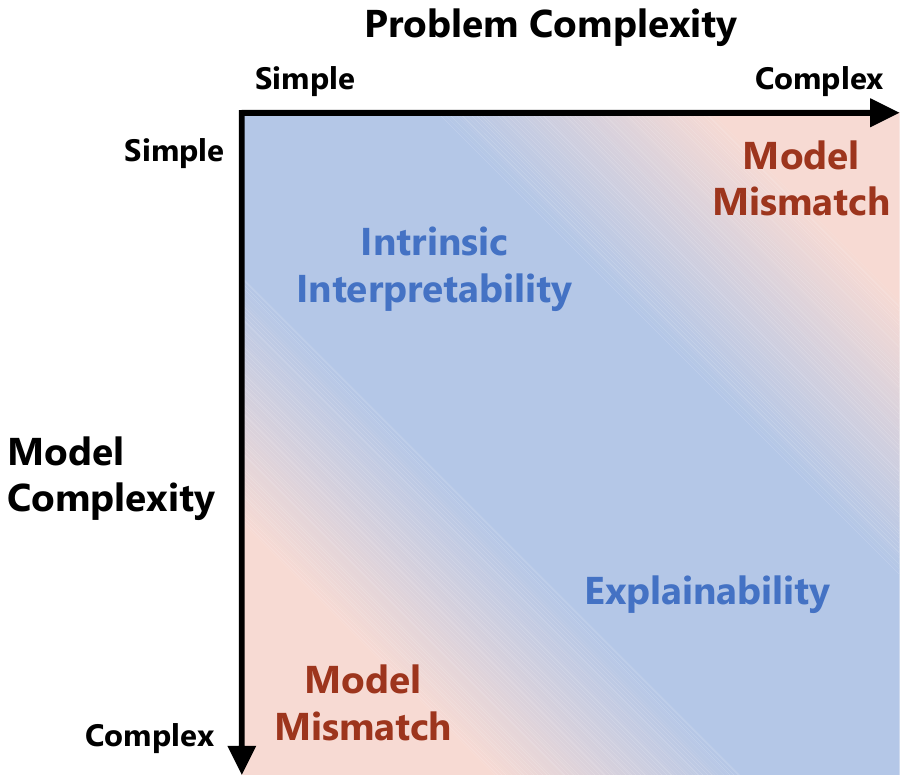}
 \caption{The interaction between problem complexity and model complexity. Simple models are inherently interpretable and suitable for simple problems, but may not capture the full behavior of complex problems. When complex models are needed for complex problems, explainability becomes essential to understand the model's behavior.}
 \label{fig:complexity}
\end{figure}

The interpretability of a model is influenced by its complexity~\cite{barceloModelInterpretabilityLens2020}. Simpler models, like linear regressions, are generally considered to be interpretable due to their straightforward decision-making processes that humans can follow unaided. An ML model is typically deemed intrinsically interpretable when it is compact and understandable. However, as model complexity grows, interpretability diminishes and we must rely on explainability.

This raises a question: why pursue explainability over creating an intrinsically interpretable model that requires no explanation? While there are reasons to favor interpretable models when possible~\cite{rudin_stop_2019}, such models are not always feasible. Moreover, even when a more complex model fits the data well, researchers warn against assuming that the reason for this is that there are underlying interpretable rules which the model has learned~\cite{HASSON2020416}. In such cases, post-hoc explanations provide the best path to understanding. They may not always be comprehensive, but provide insights that allow us to gain at least a partial understanding of complex models.

For any given problem, there is a minimum level of complexity required in the model to accurately model the data. If this level is low, a simple, interpretable model can suffice. However, when the problem demands more complexity, a complex model becomes necessary, and explainability is required. In either case, the complexity of the problem determines the complexity of the model. This relationship between problem complexity and model complexity is important for understanding when we need explainability.

To illustrate this, we present a framework in Fig.~\ref{fig:complexity}, mapping problem complexity against model complexity. The complexity of a model (or an optimization solution) includes factors such as the number of parameters, depth of structure, and computational requirements. Although we treat this informally, this may be quantified through metrics such as the model’s description length~\cite{hinton1993keeping} or parameterized complexity~\cite{barceloModelInterpretabilityLens2020}. A model with a lengthy description, numerous parameters, or complex functions is considered more complex. While increased complexity can improve problem-solving capacity, it often reduces interpretability simply due to the model’s size. For instance, neural networks with billions of parameters or deep decision trees may perform well but are difficult to interpret.

For problem complexity, we adopt an informal analogue of Kolmogorov complexity~\cite{rylander2001computational}. A problem's complexity is defined as the complexity of the simplest model required to capture its behavior at the desired level of accuracy. As problem complexity increases, so does the complexity of the model needed to represent it. Some problems may appear simple if we are satisfied with an approximation, but become complex when aiming for greater accuracy. Similar to how Kolmogorov complexity can only be approximated due to the undecidability of the halting problem, this notion of problem complexity does not assume we can identify the least complex model definitively, only that it exists.

With these two axes, problem complexity and model complexity, we can identify four key scenarios:

\begin{itemize}
\item \textbf{Simple problem, simple model:} A simple model captures the desired behavior. This model is both accurate and intrinsically interpretable, eliminating the need for explainability.
\item \textbf{Simple problem, complex model:} Using a complex model for a simple problem creates a mismatch. While the model may perform well, it is unnecessarily complex and difficult to interpret. In this case, the problem is not one of explainability issue but of model selection: a simpler model which requires no explanation would suffice. However, if such a model exists but cannot be found in practice, explainability may still be used to gain understanding.
\item \textbf{Complex problem, simple model:} Applying a simple model to a complex problem results in inadequate performance. Although the model is interpretable, it fails to capture the complexity of the data to the desired degree, leading to a mismatch between the problem and the model. This can be addressed by using a more complex model (which then requires explanation), or if we are satisfied with a less accurate solution, lowering our requirements and reducing the problem to a simple one which can be solved by a simple model.
\item \textbf{Complex problem, complex model:} This is the primary area of relevance for XAI. Complex models are required to solve complex problems, but are difficult to interpret. Explainability methods are essential to help users understand these models, since the models cannot be understood on their own and cannot be made intrinsically interpretable while still solving the problem adequately.
\end{itemize}

%%%%%%%%%%%%%%%%%%%%%%%%%%%%%%%%%%%%%%%%%%%%%%%%%%%%%%%%%%%%%

\subsection{Types of explanations}

\begin{figure}[t!]
 \centering
 \includegraphics[width=0.65\linewidth,trim=50mm 50mm 55mm 50mm,clip]{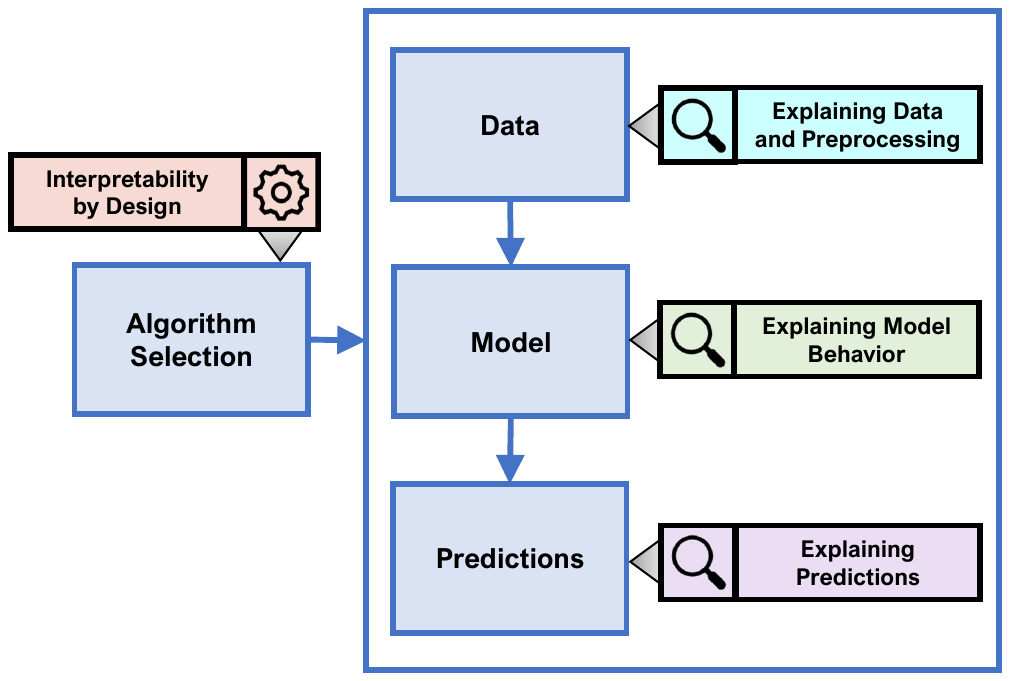}
 \caption{Overview of the process of building an ML model, showing areas where explanations (magnifying glasses) are often applied. Also shown is the intrinsic interpretability approach (cogwheel), where models are designed to be interpretable from the start. All these methods can be used together to form a complete picture of a model's behavior.}
 \label{fig:xai-ml}
\end{figure}

Explanations target various aspects of the modeling process. Here, we take a problem-focused approach, examining the entire ML pipeline, from data to trained model (Fig.~\ref{fig:xai-ml}). Our categorization is based on the stage of the ML pipeline where explainability can be applied. This approach emphasizes that explainability is not only relevant at the end of the process, once a model is fully trained, but can also enhance understanding throughout the model-building process. By aligning explanations with the pipeline stages, we aim to offer practitioners a clear roadmap for where and how explainability can be integrated.

In the following sections, we address each stage in turn. First, we introduce each category by describing examples of conventional approaches. This overview is not exhaustive but provides a primer on commonly used methods. For a more comprehensive review of current XAI methods, we refer readers to recent surveys on the topic~\cite{dwivediExplainableAIXAI2023, saeedExplainableAIXAI2023}. We then explore EC-based approaches within each stage, offering an overview of the state-of-the-art in combining EC and XAI.

%%%%%%%%%%%%%%%%%%%%%%%%%%%%%%%%%%%%%%%%%%%%%%%%%%%%%%%%%%%%%

\subsection{Interpretability by design}

As alluded to in \Cref{section:complexity}, a growing consensus in the XAI community advocates for developing interpretable ML models whenever possible, rather than relying on post-hoc explanations~\cite{iML_Rudin_NatML_19,iml_murdoch_pnas19}. The argument is that post-hoc explanations often only provide local approximations, which are limited in two key ways: 1) they rarely capture the entire decision-making process of a model, and 2) being approximations of another model, they can introduce errors, potentially distorting the original decision-making logic. As a result, models built to be inherently interpretable and grounded in knowledge representation are preferable when possible.

Unlike traditional ML models that often use greedy heuristics, EC methods leverage global optimization through evolutionary search. A notable example of this is learning classifier systems (LCS), which apply either batch~\cite{bacardit_thesis,bacardit09improving} or online learning~\cite{wilson1995classifier,holland1999learning}, using reinforcement learning (RL)~\cite{wilson1995classifier} or supervised learning~\cite{journals/ec/MansillaG03}. While most LCS approaches generate rule-based models, alternative representations such as decision trees~\cite{llora_evolution_nodate} and hyper-ellipsoids~\cite{butz_kernel_2005} have also been explored. Genetic programming (GP), as another example, also has a long history of producing interpretable, symbolic solutions to learning problems~\cite{la2021contemporary,Smith_TPG_23}.

To control model complexity, EC methods employ techniques to manage model size, such as minimizing bloat~\cite{langdon97fitness} and using fitness functions that encourage compact rule sets based on principles like minimum description length (MDL)~\cite{iwlcs2003-new}. Multi-objective optimization~\cite{llora2003accuracy} and rule-editing operators~\cite{bacardit09memetic,franco2012post} are also used to simplify models. Sparsity in neural networks, for example, can be achieved through regularization or evolutionary pruning~\cite{shangNeuralNetworkPruning2022,zhouEvolvingBetterInitializations2023}.

Efforts to combine RL and EC have aimed to produce interpretable policies by using decision trees induced by GP or grammatical evolution, with RL guiding actions~\cite{hein_interpretable_2018,custode2020evolutionary}. Studies also explore quality-diversity methods~\cite{ferigo2023qualitya,ferigo2023qualityb} and multi-agent settings~\cite{crespi2023population}. The balance between accuracy and interpretability has also been studied in genetic fuzzy systems~\cite{galende2009accuracy}, while recent work has proposed machine-learned measures of interpretability~\cite{virgolin2021model} and emphasized the importance of low-complexity models, particularly in GP~\cite{virgolin2022less}.

Visualization techniques, such as heatmaps, have been effective for interpreting classification rules generated by LCS~\cite{urbanowicz2012analysis}, particularly when combined with hierarchical clustering. Similarly, 3D visualizations have proven useful for representing rule sets, illustrating attributes, rule generality, and estimated attribute importance~\cite{liu2021visualizations}.

%%%%%%%%%%%%%%%%%%%%%%%%%%%%%%%%%%%%%%%%%%%%%%%%%%%%%%%%%%%%%

\subsection{Explaining data and preprocessing}
\label{section:explaining_data}

We begin by discussing methods aimed at explaining the data, focusing on understanding its structure before modeling. These techniques, although not applied to the final model, are nonetheless crucial to the ML pipeline. Every model begins with data, and any patterns the model learns are derived from this data. While these methods do not explain the model itself, they provide insight into the data distribution and characteristics that shape model learning.

Exploratory analysis, data visualization, and dimensionality reduction techniques, such as principal component analysis (PCA)~\cite{pearsonLIIILinesPlanes1901, hotellingAnalysisComplexStatistical1933} and t-distributed stochastic neighbor embedding (t-SNE)~\cite{maatenVisualizingDataUsing2008}, help uncover patterns and potential biases in the data. These methods simplify data for easier interpretation and visualization.

Additionally, clustering and outlier detection techniques, like k-means and DBSCAN~\cite{ester1996density}, identify patterns or anomalies that can affect model performance and inform feature selection. These explanations can help identify data quality issues, biases, and preprocessing requirements.

%%%%%%%%%%%%%%%%%%%%%%%%%%%%%%%%%%%%%%%%%%%%%%%%%%%%%%%%%%%%%

\subsubsection{Dimensionality reduction}

EC can aid in explaining data through dimensionality reduction and visualization. One approach is GP-tSNE~\cite{GPinterpVis_Lensen_ITC_21}, which adapts the t-SNE~\cite{maatenVisualizingDataUsing2008} algorithm by using evolved trees to create an interpretable mapping from original data points to embedded points. Similarly, a study~\cite{schofieldUsingGeneticProgramming2021} uses tree-based GP to generate an interpretable mapping for uniform manifold approximation and projection (UMAP)~\cite{mcinnesUMAPUniformManifold2018}. These explicit mapping functions make the process more understandable and reusable for new data.

In some cases, lower-dimensional representations are useful for both prediction and visualization. This enhances interpretability, allowing us to visualize the exact data representation seen by the model. For example, a multi-objective GP algorithm~\cite{ickeMultiobjectiveGeneticProgramming2011} is developed to optimize features for classifiability, visual interpretability, and semantic interpretability. Another study optimizes features for both visualization and downstream tasks, balancing classification metrics (accuracy, AUC, and Cohen’s kappa) with visualization metrics (C-index, Davies-Bouldin, and Dunn’s index) to improve clustering and separability~\cite{canoMultiobjectiveGeneticProgramming2017}.

GP has also been applied to manifold learning~\cite{lensen2021genetic}, creating reduced representations for high-dimensional datasets. While traditional black-box algorithms often lack transparency in how they map data to a reduced space, GP trees offer interpretable, white-box alternatives for these transformations.

%%%%%%%%%%%%%%%%%%%%%%%%%%%%%%%%%%%%%%%%%%%%%%%%%%%%%%%%%%%%%

\subsubsection{Feature selection and feature engineering}
\label{sec:feature-sel-feature-eng}

Feature selection is a common preprocessing step that selects a relevant subset of features from the original dataset. This improves both model performance and interpretability by limiting the features a model can rely on. While similar to feature importance, which identifies key features, feature selection explicitly restricts the model to the chosen subset.

Genetic algorithms are a natural and effective approach to feature selection (since solutions can represented as binary strings), and are widely used in this area~\cite{sayedNestedGeneticAlgorithm2019, shaFeatureSelectionPolygenic2021,xueMultiObjectiveFeatureSelection2022}. GP is also often used, as feature selection is inherently part of the evolved program structure~\cite{GPinterpret_Hu_GPTP_20,EvoNetOA_Hu_PLOSCB_18,SMILE_Sha_BMCBio_21}. For a detailed review of GP methods in feature selection, see~\cite{xueSurveyEvolutionaryComputation2016}. Swarm intelligence methods, such as particle swarm optimization, are another effective technique~\cite{xueSelfadaptiveParticleSwarm2019}, with further discussion available in~\cite{nguyenSurveySwarmIntelligence2020}. In addition to model improvement, feature selection aids in understanding data when combined with clustering techniques~\cite{hancerSurveyFeatureSelection2020}.

Feature engineering, or feature construction, creates higher-level features from basic ones. GP is well-suited for evolving these features for tasks like classification and regression~\cite{lacavaLearningFeatureSpaces2020,GP-FE_Li_BIBM_20, muharramEvolutionaryConstructiveInduction2005, virgolinExplainingMachineLearning2020}. This process can enhance interpretability by reducing the number of low-level features into more meaningful, higher-level ones that are easier to understand.

Feature engineering shares similarities with dimensionality reduction and, in some cases, overlaps with it. A study on various multi-tree GP algorithms for dimensionality reduction~\cite{uriotGeneticProgrammingRepresentations2022} demonstrates that GP-based methods perform comparably to traditional techniques.% like PCA, LLE, and Isomap.

%%%%%%%%%%%%%%%%%%%%%%%%%%%%%%%%%%%%%%%%%%%%%%%%%%%%%%%%%%%%%

\subsection{Explaining model behavior}

Once a model is trained, understanding how it functions can still be challenging, even if we have access to its internal mechanisms. For example, being able to examine the weights of a neural network does not necessarily make the overall model more interpretable. In these cases, explanations help bridge the gap by providing insights into how the model operates. These methods aim to explain the model’s internal structure either at the global level or for specific subcomponents.

%%%%%%%%%%%%%%%%%%%%%%%%%%%%%%%%%%%%%%%%%%%%%%%%%%%%%%%%%%%%%

\subsubsection{Feature importance}

Global feature importance explains a model’s dependence on each feature by assigning a score that reflects the significance of each feature to the model’s predictions. This helps identify which features the model is using and whether it aligns with human expectations. This technique can also aid in model optimization and feature selection by highlighting less important features. Some models, like decision trees and random forests, offer built-in feature importance measures~\cite{randforest_breiman_ml_01}, while others use more general methods like partial dependence plots~\cite{friedman1991multivariate} and permutation feature importance~\cite{fisher2019all}. EC can further refine feature importance by measuring interactions between features, evolving groups that reveal higher-order interactions~\cite{robertsonBioInspiredFrameworkMachine2022}.

% For instance, in image classification, feature importance can highlight the pixels most important to the decision and can reveal if a model is using meaningful patterns or spurious background details. 
%%%%%%%%%%%%%%%%%%%%%%%%%%%%%%%%%%%%%%%%%%%%%%%%%%%%%%%%%%%%%

\subsubsection{Global model approximations}

Global model approximations, also known as model extraction or global surrogates, approximate a black-box model with a simpler, interpretable one. This concept is related to knowledge distillation in deep learning~\cite{buciluaModelCompression2006,hintonDistillingKnowledgeNeural2015}, but with the added goal of enhancing interpretability. One approach to this is the use of interpretable decision sets to approximate a model’s behavior~\cite{lakkarajuInterpretableExplorableApproximations2017}. GP is well-suited for model extraction~\cite{iMLGP_Evans_GECCO_19}, as it can evolve decision trees that replicate the predictions of a black-box model while minimizing complexity. This method preserves accuracy while producing more interpretable models compared to other extraction techniques.

%%%%%%%%%%%%%%%%%%%%%%%%%%%%%%%%%%%%%%%%%%%%%%%%%%%%%%%%%%%%%

\subsubsection{Domain-specific knowledge extraction}

In specific domains, EC has been used to extract meaningful knowledge from ML models. For example, classification rules evolved by EC methods have been applied to protein structure prediction~\cite{bacardit_coordination_2006}. Similarly, EC-based techniques have been used to infer biological functional networks~\cite{lazzarini2016functional}, leading to experimentally validated gene discoveries in plants~\cite{bassel_functional_2011}. Knowledge representations in rule-based systems can also shape the patterns captured, resulting in different insights from the same data, as demonstrated in molecular biology~\cite{baron2017characterising}. In neuro-evolution, EC has been applied to discover interpretable plasticity rules~\cite{mettler2021evolving,jordan2020evolving}, as well as self-interpretable agents that rely on selective inputs~\cite{tang_neuroevolution_2020}.

%%%%%%%%%%%%%%%%%%%%%%%%%%%%%%%%%%%%%%%%%%%%%%%%%%%%%%%%%%%%%

\subsubsection{Explaining neural networks}

While we have focused thus far on methods which apply to a variety of models, deep learning requires specific techniques due to the complexity and black-box nature of the models. Explaining these models is exceptionally difficult due to the large number of parameters.

In image classification, the large number of input features (pixels) poses a problem for many explanation methods. Dimensionality reduction techniques, like clustering pixels into ``superpixels,” help identify important regions for predictions. Multi-objective algorithms can optimize for minimal superpixels while maximizing model confidence~\cite{wangMultiobjectiveGeneticAlgorithm2022}.

Efforts to explain internal representations~\cite{samekExplainingDeepNeural2021} include methods that create invertible mappings from complex latent spaces to simpler, interpretable ones, offering insight into how models process information~\cite{adelDiscoveringInterpretableRepresentations2018}.

Recently, ``mechanistic interpretability” has gained attention, aiming to reverse-engineer neural networks by analyzing activation patterns to reveal the underlying algorithms~\cite{cammarata2020thread,elhage2021mathematical}. This approach has explained phenomena like ``grokking,” where models initially memorize data but later generalize after extended training~\cite{nandaProgressMeasuresGrokking2023}.

EC is well-suited for explaining neural networks, notably through the ability to construct small interpretable explanations and to prioritize exploration. EC methods have been used to map out decision boundaries and input spaces in language models~\cite{salettaGrammarbasedEvolutionaryApproach2022,bradley2024openelm}, and symbolic regression has provided interpretable mathematical expressions that describe network gradients~\cite{wetzelClosedFormInterpretationNeural2024}.

%%%%%%%%%%%%%%%%%%%%%%%%%%%%%%%%%%%%%%%%%%%%%%%%%%%%%%%%%%%%%

\subsection{Explaining predictions}

This approach focuses on explaining a specific prediction rather than the model's overall behavior. The explanation only needs to capture how the model arrived at the particular prediction in question.

%%%%%%%%%%%%%%%%%%%%%%%%%%%%%%%%%%%%%%%%%%%%%%%%%%%%%%%%%%%%%

\subsubsection{Local explanations}

Local explanation approaches aim to approximate a model’s behavior for a specific prediction rather than explaining the model globally. One widely used method, local interpretable model-agnostic explanations (LIME)~\cite{expPredict_Ribeiro_KDD_16}, generates data points near the input and fits a linear model to approximate the black-box model’s behavior in that local region. While this approach doesn't capture the model's global performance, it provides a locally faithful explanation for one data point.

An alternative approach, genetic programming explainer (GPX)~\cite{GP-localExp_Ferreira_CEC_20}, uses GP to evolve symbolic expressions that better capture local patterns than LIME's linear approximation. GPX constructs a local explanation by sampling neighboring data points and evolving a model that reflects the behavior of the black-box model more effectively, particularly when linearity assumptions do not hold.

Another example, local rule-based explanations (LORE)~\cite{guidottiLocalRuleBasedExplanations2018}, introduces an evolutionary algorithm to generate a neighborhood of points around the prediction. These points are classified either similarly or differently from the original prediction, and a decision tree is used to capture the local behavior. The evolutionary algorithm ensures a dense and diverse set of neighborhood points, allowing the decision tree to provide a more robust local explanation.

In addition to approximating a local model, Shapley additive explanations (SHAP)~\cite{SHAP_Lundberg_NeurIPS_17} offers a way to assess feature importance for individual predictions by approximating Shapley values from game theory. These values represent each feature's contribution to the prediction, with SHAP using efficient sampling techniques to make this computation feasible for practical use.

%%%%%%%%%%%%%%%%%%%%%%%%%%%%%%%%%%%%%%%%%%%%%%%%%%%%%%%%%%%%%

\subsubsection{Counterfactuals}

Counterfactual explanations provide insight by offering hypothetical scenarios where the model would make a different decision. For example, ``the model would have approved the loan if the income were \$5000 higher" explains how a change in input affects the model’s output~\cite{wachterCounterfactualExplanationsOpening2017b}. These explanations are intuitive, model-agnostic, and provide users with actionable steps, or recourse, to achieve desired outcomes~\cite{karimiSurveyAlgorithmicRecourse2020}. However, since they focus on single instances, they offer limited insight into the model’s global behavior.

Diverse counterfactual explanations (DiCE)~\cite{mothilalExplainingMachineLearning2020} generates counterfactuals that are valid (produce a different outcome), proximal (similar to the original input), and diverse (distinct from each other). Diversity enhances the usefulness of explanations by offering multiple perspectives on how the model behaves. DiCE achieves this using determinantal point processes~\cite{kuleszaDeterminantalPointProcesses2012}, which balance proximity and diversity while ensuring that only a few features differ from the original input.

EC is well suited for generating counterfactuals thanks to its black-box optimization capabilities and ability to handle multiple objectives. CERTIFAI~\cite{sharmaCERTIFAICommonFramework2020} applies a genetic algorithm to generate counterfactual explanations by sampling instances on the opposite side of the model's decision boundary. The genetic algorithm then optimizes the population of counterfactuals by minimizing their distance from the original input instance. This approach measures robustness (based on how far counterfactuals are from the input) and fairness (comparing robustness across different feature values). GeCo~\cite{schleichGeCoQualityCounterfactual2021} also uses a genetic algorithm, but with additional constraints to ensure plausibility, such as avoiding unrealistic changes (e.g., altering age or gender). The algorithm prioritizes proximity to the decision boundary and minimizes feature changes to simplify the explanation. Multi-objective counterfactuals (MOC)~\cite{dandlMultiObjectiveCounterfactualExplanations2020} explicitly optimizes for multiple criteria using a modified NSGA-II algorithm. MOC balances four objectives: achieving the desired model output, maintaining proximity to the input, minimizing feature changes, and ensuring plausibility (measured by distance to real data points). The use of mixed integer evolution strategies (MIES)~\cite{liMixedIntegerEvolution2013} allows MOC to search both discrete and continuous spaces efficiently.

%%%%%%%%%%%%%%%%%%%%%%%%%%%%%%%%%%%%%%%%%%%%%%%%%%%%%%%%%%%%%

\subsubsection{Adversarial examples}

Adversarial examples are closely related to counterfactuals but are designed to intentionally produce incorrect predictions~\cite{IML_eBook_Molnar_22}. These examples are created by applying small perturbations to inputs, changing their classification while keeping them perceptually similar to the original. Adversarial examples highlight failure modes in models and serve as potential attack vectors, especially for deep learning systems.

One approach~\cite{suOnePixelAttack2019} generates adversarial examples by modifying just a single pixel in an image. This contrasts with previous methods that altered multiple pixels and were more noticeable. Using differential evolution, the method optimizes the pixel coordinates and perturbation in RGB space, demonstrating that a single pixel is often enough to fool the model. Other works have explored evolution strategies~\cite{qiu2021black}, differential evolution~\cite{liBayesianEvolutionaryOptimization2023}, multi-objective evolutionary algorithms~\cite{suzukiAdversarialExampleGeneration2019}, and the clonal selection algorithm~\cite{custode2021one} to generate adversarial image perturbations.

Adversarial examples are not limited to image models. In natural language processing, adversarial examples have been generated for sentiment analysis and textual entailment models~\cite{alzantotGeneratingNaturalLanguage2018}. In this case, adversarial inputs are semantically and syntactically similar to the original, making the attack harder to detect. A genetic algorithm optimizes the input for a target label, with mutations changing words to their nearest neighbors in a word embedding model and removing words to ensure the context remains intact.

%%%%%%%%%%%%%%%%%%%%%%%%%%%%%%%%%%%%%%%%%%%%%%%%%%%%%%%%%%%%%

\subsection{Assessing explanations}
\label{section:assessing_explanation}

In addition to generating explanations, EC can be used to assess or enhance the quality of other explanation methods.
One study~\cite{huangSAFARIVersatileEfficient2022} proposes two metrics to evaluate the robustness of an explanation: worst-case misinterpretation discrepancy and probabilistic interpretation robustness. Interpretation discrepancy measures the difference between two interpretations, before and after input perturbation. A robust interpretation should have a low discrepancy. A genetic algorithm is used to optimize two worst-case scenarios: the largest discrepancy when the classification remains unchanged, and the smallest discrepancy when the classification changes (adversarial example). The second metric, probabilistic misinterpretation, estimates the likelihood of significant interpretation changes under these conditions using subset simulation.

EC can also be applied to adversarial attacks on explanations themselves. One such method, AttaXAI~\cite{tamamFoilingExplanationsDeep2022}, evolves images that appear similar to the original input with the same model prediction but an arbitrary explanation map. Experiments show that, using pairs of images, AttaXAI can create a new image with the same appearance and prediction as the first image, while the explanation map resembles that of the second.

%%%%%%%%%%%%%%%%%%%%%%%%%%%%%%%%%%%%%%%%%%%%%%%%%%%%%%%%%%%%%

\section{XAI for EC}
\label{sec:xai_for_ec}

\begin{figure}[t!]
 \centering
 \includegraphics[width=0.6\linewidth,trim=20mm 50mm 90mm 30mm,clip]{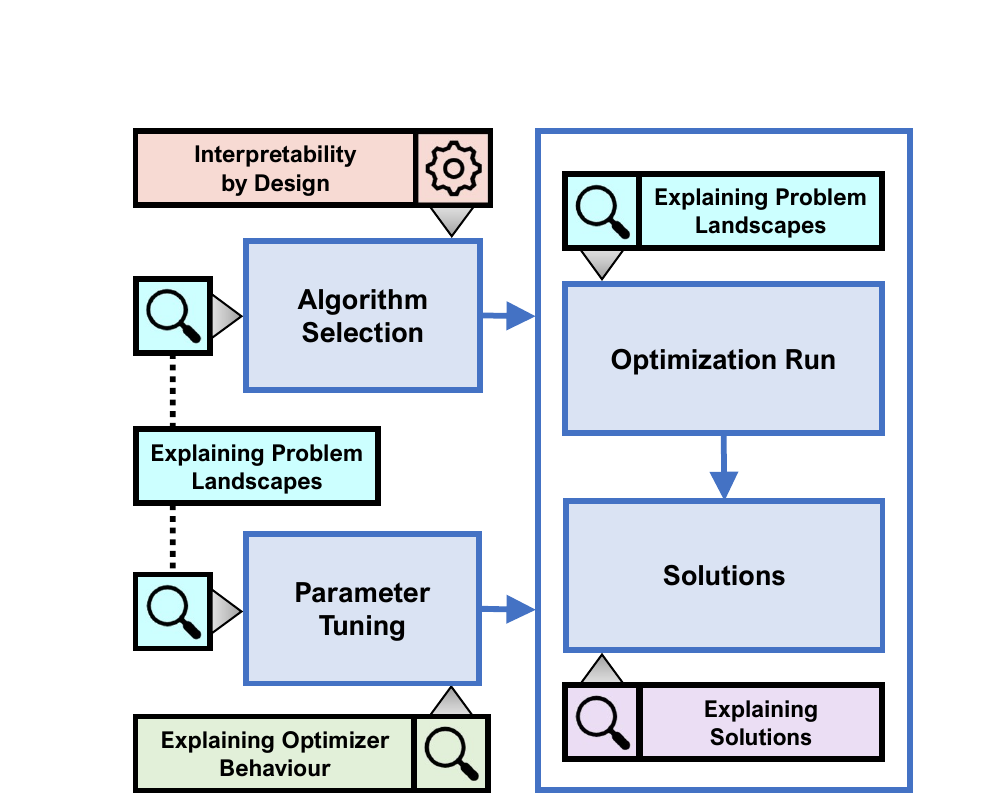}
 \caption{Overview of the process of using an optimization algorithm, 
 showing areas where explanations (magnifying glasses) can be applied.}
 \label{fig:xai-ec}
\end{figure}

In this section, we shift the focus to explainability for EC and optimization methods in general. Similar to the previous discussion, optimization algorithms often involve lengthy, complex processes to find optimal or near-optimal solutions for decision-makers. Explanations in this context help answer the overarching questions introduced in \Cref{section:xai}. We view the optimization process (Fig.~\ref{fig:xai-ec}) as comprising four key stages: algorithm selection, parameter tuning, iterative search, and solution analysis. Each stage presents opportunities for explainability, which we will elaborate on in the following sections.

%%%%%%%%%%%%%%%%%%%%%%%%%%%%%%%%%%%%%%%%%%%%%%%%%%%%%%%%%%%%%

\subsection{Interpretability by design}

A key challenge in optimization lies in designing and formulating objectives and solution representations. Interpretability plays a crucial role in these design choices, as the way a problem is defined directly impacts how it is explained. To improve interpretability, more direct representations and explicit encoding of variables, objectives, and constraints may be preferred. For example, a mixed-integer linear programming (MILP) formulation, whether solved by mathematical optimization or EC, is preferred than a ``black-box" function evaluation. Matheuristics~\cite{fischetti2018matheuristics} have proven successful in this area. Another approach~\cite{GOERIGK20231312} uses decision trees to provide interpretable rules for selecting solutions, with trees constructed by MILP or heuristics.

Handling components of an objective separately, rather than combining them, allows for post-hoc analysis of the evolutionary process. This motivates the use of lexicographical approaches to tournament selection, such as those proposed for constraints~\cite{deb2000efficient} and multiple objectives~\cite{deb2002fast}, or lexicase selection in GP~\cite{helmuth2014solving}. Multi-objective problems can also be made more understandable using post-hoc multi-objective evolutionary algorithms~\cite{xu2020survey,falcon2020indicator,hua2021survey} that approximate a Pareto-front, enabling decision-makers to better understand trade-offs between objectives, rather than guessing through weighted sums. This topic will be revisited in \Cref{sec:explaining-solutions}.

Direct representations are often favored for explainability, as they align more closely with real-world decision variables, making explanations easier in applied settings. However, there is a trade-off: indirect representations, such as hyperNEAT~\cite{stanley2009hypercube,d2014hyperneat} and grammatical evolution, often outperform direct representations like classical GP trees~\cite{Nyathi2018Comparison}. Direct formulations allow for greater control of operators, tailored to the problem at hand. For example, grey-box optimization~\cite{Whitley2016} leverages domain knowledge with direct encodings in combinatorial problems to improve performance. As with ML (see \Cref{sec:feature-sel-feature-eng}), representations must balance granularity and interpretability. Overly detailed representations become too dense to interpret, while high-level abstractions may lose real-world relevance.

There may also be opportunities for evolutionary algorithms to select or engineer interpretable features for optimization problems. An example on this research idea is achieved through cooperative coevolution, a technique already successful in large-scale global optimization~\cite{yang2008large}.

The choice of algorithmic framework also affects explainability. Greedy algorithms or steepest ascent hill-climbers, being deterministic, provide a single, easily traceable path, making them more interpretable than stochastic or population-based algorithms. Estimation of distribution algorithms~\cite{larranaga2001estimation,Ceberio2011,Hauschild2011a,Lozano2006} construct explicit problem representations and clear mathematical routes to solutions, although these processes can become complex and harder to interpret.

%%%%%%%%%%%%%%%%%%%%%%%%%%%%%%%%%%%%%%%%%%%%%%%%%%%%%%%%%%%%%

\subsection{Explaining problem landscapes}

Landscape analysis focuses on understanding the interactions between algorithms, their operators, and solution representations. While this approach emphasizes understanding \textit{how} the search proceeds, rather than \textit{why} specific solutions are chosen, both aspects are crucial for improving explainability in optimization.

%%%%%%%%%%%%%%%%%%%%%%%%%%%%%%%%%%%%%%%%%%%%%%%%%%%%%%%%%%%%%

\subsubsection{Landscape analysis and trajectories}

Landscape analysis~\cite{malan_survey_2021} is a key intersection between XAI and EC, offering tools to understand algorithm behavior based on problem features, predict performance, and optimize algorithm selection and configuration. Recent works have focused on explainable landscape-aware predictions~\cite{trajanov2021explainable,trajanov2022explainable}.

An algorithm’s behavior can be described by its trajectory through the search space, as the sequence of points visited during a run. This trajectory captures insightful information, such as when solutions are discovered, when the algorithm converges prematurely, or gets stuck in a local optimum. Search trajectory networks~\cite{OCHOA2021SearchTrajectoryNetworks} visualize these paths, helping to explain the algorithm’s progress for various problem-algorithm combinations.

Search trajectory analysis has also been suggested as a promising technique for XAI in EC~\cite{Fyvie2021}. By applying principal component analysis (PCA) to solutions, it becomes possible to capture dominant features in each generation, visualize algorithm progress, and relate components to known global optima. Other work~\cite{DynamoRep} proposes using simple descriptive statistics to characterize optimization trajectories, which can then be used in ML methods for performance prediction or automatic algorithm configuration.

Population dynamics plots~\cite{Walter2022Explainable} visualize EA progress and convergence behavior by tracing the lineage of solutions and their proximity to feasibility boundaries. These visualizations are especially useful in multi-objective problems like knapsack optimization, projecting multi-dimensional solutions into interpretable two-dimensional forms.

Another approach involves creating surrogate fitness models biased toward solutions visited during the evolutionary search~\cite{Wallace2021,Singh2022}. Probing these models reveals insights into variable sensitivity and inter-variable relationships, offering another lens through which to study the algorithm’s trajectory.

Research into hyper-heuristics~\cite{drake2020recent} and parameter selection~\cite{smit2010parameter} highlights certain parameter configurations that enable an evolutionary algorithm to perform well across diverse functions. Exploring whether simpler, more explainable parameter settings can also lead to generalist solutions could be a promising area of research, aligning with principles like Occam's razor.

%%%%%%%%%%%%%%%%%%%%%%%%%%%%%%%%%%%%%%%%%%%%%%%%%%%%%%%%%%%%%

\subsubsection{User-guided evolution}

Allowing users to interact with the model-building process can enhance explainability. One approach~\cite{Urquhart2017} combines parallel coordinate plots with a multi-objective EA, enabling users to define areas of interest where they want solutions.

Quality-diversity or illumination algorithms, like MAP-Elites~\cite{mouret2015illuminating,Gaier2017}, offer another way to explore solution landscapes. These algorithms generate diverse, high-quality solutions along user-defined dimensions, helping users understand how solution quality varies with respect to different parameters. A future direction could involve designing algorithms that provide human-interpretable explanations throughout the process, incorporating user feedback during the search, similar to preference-based multi-objective optimization~\cite{miettinen2000interactive}. 

The efficiency of MAP-Elites can be improved through hybrid approaches~\cite{Gaier2017} using a surrogate model and an intelligent sampling of fitness evaluation. Another application of MAP-Elites~\cite{Urquhart:2019:ITM:3319619.3326816} addresses the lack of user involvement. By filtering the solution space and offering users a set of solutions to choose from, users can gain influence over what constitutes a ``good" solution. More recently, MAP-Elites has been extended to extract explainable rules from its archives~\cite{Urquhart2021}. This work addresses the challenge of interpreting thousands of solutions by using GP and rule induction to generate a small set of rules that describe the characteristics of the solutions produced by the optimizer.

%%%%%%%%%%%%%%%%%%%%%%%%%%%%%%%%%%%%%%%%%%%%%%%%%%%%%%%%%%%%%

\subsection{Explaining solutions}
\label{sec:explaining-solutions}

The solutions generated by optimization, whether they are Pareto-fronts, populations, or single solutions, can be examined for explanatory insights. This post-hoc analysis explores alternative causes to explain solution quality and reveal underlying aspects of the model and the algorithm.

%%%%%%%%%%%%%%%%%%%%%%%%%%%%%%%%%%%%%%%%%%%%%%%%%%%%%%%%%%%%%

\subsubsection{Interpreting solutions}

Interpretability is related to the concept of \textit{backbones}~\cite{Walsh2001} in optimization, which represent critical components of a solution. For example, in a satisfiability decision problem, a backbone consists of literals that are true in every model. Identifying these features in a solution can provide an explanation for its quality.

Dimensionality reduction techniques are helpful in explaining optimizer solutions. For instance, using multiple correspondence analysis (MCA) can decompose search trajectories, projecting them into lower-dimensional spaces to highlight feature importance at different stages of the search~\cite{Fyvie_StaffRostering2}. This helps interpreting the influences that impact the quality of solutions in single-objective problems.

In multi-objective optimization, the trade-off between a solution's explainability and its accuracy has been explored~\cite{tiwonge_zav_explain_tradeoffs}. By applying step-wise regularization to linear regression models generated by an optimizer, the complexity of explanation representations is reduced while maintaining predictive ability and interpretability.

The concept of innovisation~\cite{Deb2008,Deb2014} aims to identify shared design principles in multi-objective solutions, explaining Pareto-optimality by highlighting key principles. More recently, efforts to maintain coherence between Pareto-front solutions have been explored\cite{scholman2022obtaining}, offering a smoother view of transitions in the solution space between solutions in Pareto-front approximations.

%%%%%%%%%%%%%%%%%%%%%%%%%%%%%%%%%%%%%%%%%%%%%%%%%%%%%%%%%%%%%

\subsubsection{Visualization of solutions}

In many-objective optimization, a challenge is to visualize Pareto-front approximation with more than three objectives, as human cognition struggles with comprehending higher-dimensional spaces. Visualization efforts have focused on three approaches: 1) techniques that display solutions in terms of all objectives, 2) identifying and discarding redundant objectives to allow for standard visualization methods, and 3) using feature extraction to create new, easier-to-visualize coordinate sets.

The first approach includes techniques like parallel coordinate plots~\cite{Inselberg2009, Urquhart2018} and heatmaps~\cite{Pryke2007}, which are widely used to visualize large datasets. However, both suffer from clarity issues: parallel coordinate plots can obscure solutions due to overlap, and heatmaps often have arbitrary ordering of rows and columns, making relationships between solutions and objectives difficult to interpret. Improvements such as reordering objectives to highlight trade-offs~\cite{Lygoe2013} or using clustering techniques~\cite{Pryke2007, Walker2013} have helped clarify these visualizations. Interactive features in parallel coordinate plots~\cite{Urquhart2018} also reduce cognitive load by allowing users to filter out irrelevant solutions.

The second and third approaches involve dimensionality reduction and feature extraction techniques like PCA~\cite{Jolliffe2002}, self-organizing maps (SOM)~\cite{Kohonen1995}, and multidimensional scaling (MDS)~\cite{Sammon1969}, which project objective vectors from $\mathbb{R}^{M>3}$ into lower-dimensional spaces ($\mathbb{R}^{M\in{2,3}}$). This allows for the use of standard visualization tools like scatter plots. However, these projections can disconnect decision-makers from the original objectives, potentially causing confusion. Improvements like varying color schemes based on objectives or annotating projected solutions with key information (e.g., best/worst solutions)~\cite{Walker2013} help mitigate this issue. Further improvements~\cite{Walker2014} have focused on identifying and visualizing the edges of the Pareto-front to better illustrate distances to extreme solutions in lower-dimensional spaces.

%%%%%%%%%%%%%%%%%%%%%%%%%%%%%%%%%%%%%%%%%%%%%%%%%%%%%%%%%%%%%

\subsection{Explaining optimizer behavior}

The analysis of optimizers is closely related to landscape analysis, as researchers seek to distinguish between the effects of an optimizer's internal mechanics and the influence of the search landscape. One approach uses special functions, such as the $f_0$ function~\cite{KONONOVA2015468}, a uniform random fitness function, or constant functions, to repeatedly assess behavioral patterns and observe the distribution of final solutions. Another method uses large and diverse benchmark sets or gradually alters benchmark function properties using affine combinations~\cite{vermetten2023affine,vermetten2023ma}.

Behavior-based benchmarks offer deeper insights into the dynamics of metaheuristics. The BIAS toolbox~\cite{vermetten2022bias,vermetten2022using} analyzes structural bias (SB) in optimization algorithms, which refers to intrinsic biases in iterative optimization algorithms that drive search towards certain regions of the solution space, independent of the objective function. The toolbox helps identify the presence, intensity, and nature of SB, while recent work~\cite{van2023deep} applies deep learning and XAI techniques to detect and analyze SB patterns, shedding light on algorithm improvement areas.

The concept of ``explainable benchmarking"~\cite{van2024explainable} introduces a framework and its software that dissects the performance of optimization algorithms by analyzing their components and hyper-parameters. Using TreeSHAP and other XAI techniques, the framework visualizes the contribution of each component to overall performance on various types of objective functions. Similarly, f-ANOVA~\cite{nikolikj2024quantifying} helps quantify which modular components of an algorithm contribute most to its optimization performance, providing insights into the relationship between algorithm configuration, problem landscape characteristics, and algorithm performance.

Another method for explaining algorithm behavior involves comparing historical benchmarking experiments and unifying their results using ontologies~\cite{yaman2017presenting}. The OPTION ontology~\cite{kostovska2021option} provides a semantic vocabulary for annotating algorithms, problems, and evaluation metrics, improving interoperability and reasoning. Recent work~\cite{Kostovska2023} extends OPTION to represent modular black-box optimization algorithms and builds knowledge graphs to predict performance based on modular algorithm configurations, leading to explainable predictions.

%%%%%%%%%%%%%%%%%%%%%%%%%%%%%%%%%%%%%%%%%%%%%%%%%%%%%%%%%%%%%

\section{Research Outlook}
\label{section:outlook}

The works discussed above are not exhaustive, and as the field of XAI continues to grow rapidly, we anticipate more studies exploring the intersection of EC and XAI. In particular, we expect an increasing number of hybrid systems that combine EC-induced interpretable models with black-box models for feature extraction and data manipulation. Such combinations could leverage the strengths of both approaches, fully exploiting the exploration capabilities unique to EC.

%%%%%%%%%%%%%%%%%%%%%%%%%%%%%%%%%%%%%%%%%%%%%%%%%%%%%%%%%%%%%

\subsection{Challenges}

One of the main challenges for evolutionary approaches to XAI, and for XAI in general, is scalability. As data and ML models become more complex, the number of parameters and features to be optimized increase as well. Methods that work well on small models and datasets may become computationally expensive when scaled up on larger ones. Yet, large models are the most incomprehensible and in need of explanation, making scalability crucial for applying XAI methods to more complex models. In particular, producing fully interpretable global explanations that accurately capture model behavior while being simple enough to understand becomes more challenging as models scale up, necessitating local explanations or a focus on explaining specific properties or components of the model. 

EC offers a promising automated approach to explainability, using evolutionary search to find local explanations and to optimize specific properties. While counterfactual examples have explored this concept, it could be extended to other explanation types. Evolutionary ML has proposed various scaling mechanisms~\cite{bacardit2013large} that could be adapted for EC-based XAI methods.

Another challenge is incorporating domain knowledge into XAI. Current XAI methods are typically broad and problem-agnostic, but integrating subject matter expertise or prior knowledge can improve explanation quality. For example, evaluating how well a genomics model aligns with existing gene associations can help validate its results. Domain knowledge can be incorporated through expert rules, constraints, or structured data like graphs. It can also improve interpretability by constraining models to focus on plausible associations or exclude irrelevant features. 

EC methods are well-suited for leveraging domain knowledge for building better models due to 1) their global search capabilities with robust and complex optimization, 2) their potential for hybridization with local search mechanisms tailored to exploit domain knowledge, and 3) their flexibility in exploration mechanisms to use domain knowledge.

%%%%%%%%%%%%%%%%%%%%%%%%%%%%%%%%%%%%%%%%%%%%%%%%%%%%%%%%%%%%%
\subsection{Opportunities}

We see several opportunities for future research using EC for XAI. One promising direction is the use of multiple objectives to optimize explanations. Explainability is inherently a multi-objective problem, as explanations must be faithful to the ML model while remaining simple enough to be interpretable. EC is well-suited for this, offering a framework for balancing these objectives. Incorporating multi-objective optimization into explanation methods could significantly enhance explanation quality.

The strength of EC on searching for diverse and novel solutions also presents opportunities for improving explanations. Quality-diversity algorithms can generate a range of explanations that offer different perspectives on a model's behavior. For example, applying this approach to counterfactual explanations could showcase a variety of model behaviors. Previous work~\cite{urquhart2021automated,walton2024does,10287177} has demonstrated the explanatory value of search space illumination, but there remains potential for new methods that better interpret and analyze solution sets to support decision-making. 

Incorporating user feedback is another promising direction, for both EC and XAI. Explanations are meant for human users, making their quality subjective and dependent on individual preferences. By integrating user feedback into the evolutionary process, explanations can become more tailored and continuously improved. Additionally, developing better metrics for evaluating explanation quality is essential to avoid overwhelming users. Future research could explore designing new operators and algorithms that explicitly generate explanations as part of the search process. 

Finally, innovations in visualization, interactivity, and sensitivity analysis will further enrich both EC and XAI. Advanced visualization techniques can help users better understand complex solution spaces and relationships between variables. Interactive tools that let users adjust parameters offer deeper insights into model behavior, while sensitivity analysis reveals how changes in inputs affect outcomes. Together, these methods improve user understanding by highlighting key features and making explanations more personalized and intuitive.

%%%%%%%%%%%%%%%%%%%%%%%%%%%%%%%%%%%%%%%%%%%%%%%%%%%%%%%%%%%%%

\subsection{Real-world impacts}

As AI becomes increasingly integrated into real-world applications, developing effective AI explanation methods is critical. It is equally important to consider the practical benefits that XAI research can bring. Here, we highlight a few application areas where evolutionary approaches to XAI can have a significant impact.

Healthcare is an especially high-stake domain. Without explanations, incomprehensible models may be ignored by clinicians, wasting resources, while flawed models may cause harm to patients. Even models with few errors may exhibit systematic biases, such as underdiagnosing certain patient groups~\cite{seyyed-kalantariUnderdiagnosisBiasArtificial2021}. Explainability can help identify these errors and biases~\cite{arias-duartFocusRatingXAI2022}. 

In the financial sector, AI models are widely used for fraud detection and risk assessment. Systematic biases in these models can also be harmful, such as disproportionately denying loans to certain groups. Additionally, regulatory requirements often mandate explainability of AI systems to ensure compliance and transparency.

Explainability has the potential to drive advancements in engineering and scientific discovery. AI is used in fields like materials design, drug discovery, and genomics, where explanations can uncover underlying mechanisms, support hypothesis generation, and validate domain knowledge.

In natural language processing, foundation models, generalist deep learning models trained on vast datasets and fine-tuned for specific tasks, have advanced rapidly and been deployed widely~\cite{bommasaniOpportunitiesRisksFoundation2021}. These models can perform tasks beyond what they are specifically trained for, but how they make decisions or generate outputs remain unclear. Any errors or biases in these models may be propagated to application-specific models built on top of them. As foundation models become more pervasive, understanding their behavior and identifying failure modes will be crucial for ensuring their reliability in applications.

%%%%%%%%%%%%%%%%%%%%%%%%%%%%%%%%%%%%%%%%%%%%%%%%%%%%%%%%%%%%%
\section{Conclusion}
\label{sec:conclusion}

We have demonstrated a strong mutual connection between EC and XAI. However, several research opportunities remain under-explored, including: 1) developing tools,  whether analytical, visual, data-driven, or model-based, to explain EC methods, their internal functioning, results, and properties/settings/instances that make an algorithm suitable for a given task; 2) defining how EC solutions should be verified and the level of problem knowledge needed to interpret them;  and 3) leveraging EC’s strengths to provide effective explanations or to evolve interpretable models by design. Another key area for exploration is the relationship between XAI and neuroevolution or neural architecture search, particularly regarding the link between optimized architectures and explainability (e.g., smaller networks may be easier to explain).

As XAI continues to grow, its importance for AI cannot be overstated. With the increasing deployment of ML and optimization systems in real-world applications, understanding these intelligent systems and their learned behaviors is more critical than ever. EC is well-positioned to contribute. In this paper, we explored various paradigms for explaining ML models and how EC can fit within these frameworks. EC excels at optimizing difficult interpretability metrics, handling non-differentiability, population-based diversity, and multi-objective optimization, offering distinct advantages for XAI. While some methods have begun leveraging these strengths, much more remains to be explored.

\bibliographystyle{unsrtnat}
\bibliography{main.bib}  %%% Uncomment this line and comment out the ``thebibliography'' section below to use the external .bib file (using bibtex) .





% \end{thebibliography}

\end{document}